# Quantifying intra-tumoral genetic heterogeneity of glioblastoma toward precision medicine using MRI and a data-inclusive machine learning algorithm

Lujia Wang[1], Hairong Wang[1], Fulvio D'Angelo[2], Lee Curtin[3], Christopher P. Sereduk[3], Gustavo De Leon[3], Kyle W. Singleton[3], Javier Urcuyo[3], Andrea Hawkins-Daarud[3], Pamela R. Jackson[3], Chandan Krishna[3], Richard S. Zimmerman[3], Devi P. Patra[3], Bernard R. Bendok[3], Kris A. Smith[4], Peter Nakaji[4], Kliment Donev[5], Leslie C. Baxter[6], Maciej M. Mrugała[7], Michele Ceccarelli[8], Antonio Iavarone[2], Kristin R. Swanson[3], Nhan L. Tran[3,9], Leland S. Hu[10¶], Jing Li[1¶*]

[1] H. Milton Stewart School of Industrial and Systems Engineering, Georgia Institute of Technology, Atlanta, Georgia, USA

[2] Institute for Cancer Genetics, Columbia University Medical Center, New York City, New York, USA

[3] Department of Neurosurgery, Mayo Clinic Arizona, Phoenix, Arizona, USA

[4] Department of Neurosurgery, Barrow Neurological Institute - St. Joseph's Hospital and Medical Center, Phoenix, Arizona, USA

[5] Department of Pathology, Mayo Clinic Arizona, Phoenix, Arizona, USA

[6] Department of Neuropsychology, Mayo Clinic Arizona, Phoenix, Arizona, USA

[7] Department of Neuro-Oncology, Mayo Clinic Arizona, Phoenix, Arizona, USA

[8] Department of Electrical Engineering and Information Technology, University of Naples "Federico II", Naples, Italy

[9] Department of Cancer Biology, Mayo Clinic Arizona, Phoenix, Arizona, USA

[10] Department of Radiology, Mayo Clinic Arizona, Phoenix, Arizona, USA

*Corresponding author

E-mail: jing.li@isye.gatch.edu (JL)

¶These authors are Joint Senior Authors.

# Abstract

*Background and Objective:* Glioblastoma (GBM) is one of the most aggressive and lethal human cancers. Intra-tumoral genetic heterogeneity poses a significant challenge for treatment. Biopsy is invasive, which motivates the development of non-invasive, MRI-based machine learning (ML) models to quantify intra-tumoral genetic heterogeneity for each patient. This capability holds great promise for enabling better therapeutic selection to improve patient outcome.

*Methods:* We proposed a novel Weakly Supervised Ordinal Support Vector Machine (WSO-SVM) to predict regional genetic alteration status within each GBM tumor using MRI. WSO-SVM was applied to a unique dataset of 318 image-localized biopsies with spatially matched multiparametric MRI from 74 GBM patients. The model was trained to predict the regional genetic alteration of three GBM driver genes (EGFR, PDGFRA and PTEN) based on features extracted from the corresponding region of five MRI contrast images. For comparison, a variety of existing ML algorithms were also applied. Classification accuracy of each gene were compared between the different algorithms. The SHapley Additive exPlanations (SHAP) method was further applied to compute contribution scores of different contrast images. Finally, the trained WSO-SVM was used to generate prediction maps within the tumoral area of each patient to help visualize the intra-tumoral genetic heterogeneity.

*Results:* WSO-SVM achieved 0.80 accuracy, 0.79 sensitivity, and 0.81 specificity for classifying EGFR; 0.71 accuracy, 0.70 sensitivity, and 0.72 specificity for classifying PDGFRA; 0.80 accuracy, 0.78 sensitivity, and 0.83 specificity for classifying PTEN; these results significantly outperformed the existing ML algorithms. Using SHAP, we found that the relative contributions of the five contrast images differ between genes, which are consistent with findings in the literature. The prediction maps revealed extensive intra-tumoral region-to-region heterogeneity within each individual tumor in terms of the alteration status of the three genes.

*Conclusions:* This study demonstrated the feasibility of using MRI and WSO-SVM to enable non-invasive prediction of intra-tumoral regional genetic alteration for each GBM patient, which can inform future adaptive therapies for individualized oncology.

**Keywords**: machine learning, brain cancer, imaging genetics

# 1. Introduction

Glioblastoma (GBM) is one of the most aggressive and lethal human cancers, with a median overall survival of only about 15 months despite best available standard therapy [1]. Intra-tumoral genetic heterogeneity is a major contributor to poor clinical outcomes [2]. Each tumor is comprised of genetically distinct subpopulations with different sensitivities to treatment, and genetic targets from one biopsy location may not accurately reflect those from other parts of the same tumor [3]. Moreover, due to the invasive nature of the disease, diffusely invaded GBM cells are always left behind in the brain after resection, and these residual regions may be genetically distinct from the biopsy samples collected during surgery [4,5]. The region-to-region genetic variability within a single tumor provides potential mechanisms for therapeutic escape and makes single targeted therapies less effective [6].

There are substantial challenges for quantifying intra-tumoral genetic heterogeneity of GBM. Ideally, one would want to take biopsy samples from many different regions of a tumor and perform genetic analysis of each sample. This, however, is infeasible due to the invasive nature of biopsy. Although the central tumor mass can often be surgically removed, the invasive portions of the tumor are often left unresected and unbiopsied given the risk to adjacent neurologic structures. Thus, biopsy alone is insufficient to characterize the full landscape of the intra-tumoral heterogeneity [2][7].

Neuroimaging techniques, such as MRI, provide data of the entire tumor and even the whole brain in a non-invasive manner. The emerging field of radiogenomics has demonstrated the feasibility of using MRI features to predict genetic characteristics of GBM via machine learning (ML). For example, Kha *et al.* [8] proposed an eXtreme Gradient Boosting (XGBoost)-based model to predict the 1p/19q codeletion status in a binary classification task for lower-grade gliomas. Lam *et al.* [9] developed a hybrid machine learning-based radiomics by incorporating a genetic algorithm and XGBoost classifier to classify low-grade glioma molecular subtypes. Akbari *et al.* [10] used a Support Vector Machine (SVM) to predict Epidermal Growth Factor Receptor (EGFR)-vIII mutation based on multiparametric MRI features extracted from tumor regions. Tykocinski *et al.* [11] predicted EGFR-vIII mutation based on features extracted from perfusion-weighted MRI using multivariable logistic regression. KickingeredeThe tir *et al.* [12] utilized stochastic gradient boosting machine, random forest, and logistic regression to predict the copy number variants (CNVs) of several GBM driver genes such as EGFR, Platelet-Derived Growth Factor Receptor Alpha (PDGFRA), and Phosphatase and Tensin Homolog (PTEN) based on multiparametric MRI. Chen *et al.* [13] developed a convolutional neural network to predict PTEN mutation using multiparametric MRI. However, these existing studies focus on predicting overall or average genetic status for the entire tumor, so they are suitable for relatively homogeneous tumors where genetic status does not significantly vary region-to-region. Although these studies have demonstrated the predictive utility of MRI, they fall short for identifying intra-tumoral or regional genetic heterogeneity within each tumor.

This paper aims to develop an ML model that can predict the genetic status region-by-region within a tumoral area of interest (AOI) of each patient using MRI. The model, denoted as $f: x \rightarrow y$, takes as input a vector $x$ consisting of MRI features extracted from each region within a tumoral AOI and outputs the genetic status of that region, $y$, where $y = 1$ or $2$ represents that the gene is non-altered or altered, respectively. The resulting regional predictions can be used to generate a prediction map that reveals the intra-tumoral heterogeneity across the AOI.

To train the ML model $f$, a binary classification approach can be considered by using a training set consisting of $(x_i, y_i)$ for $n$ biopsy samples. However, the biopsy sample size is often small, and a more robust approach is to use semi-supervised learning (SSL) [14]. SSL trains $f$ by including both the biopsy/labeled samples $(x_i, y_i)$ and unlabeled tumoral samples, $(x_j)$, i.e., samples from the unbiopsied regions of the tumoral AOI. Additionally, it is possible to leverage samples from outside the tumoral AOI (i.e., the normal brain area), $(x_k)$. To include these normal brain samples, one option is to treat them as a third class (class 0), in addition to non-altered gene (class 1) and altered gene (class 2) within the tumoral AOI, and train a three-class classifier. The other option, which may be more appropriate, is to train an ordinal classifier [15–17] by considering that class 0, 1, and 2 have an intrinsic order of increasing abnormality. Fig 1 illustrates the different modeling options. However, none of these models can include all available data. To address this gap, we propose a new model called Weakly-Supervised Ordinal SVM (WSO-SVM), which is designed to integrate unlabeled tumoral samples and normal brain samples beyond just biopsy (labeled) samples to enhance the model's learning capacity.

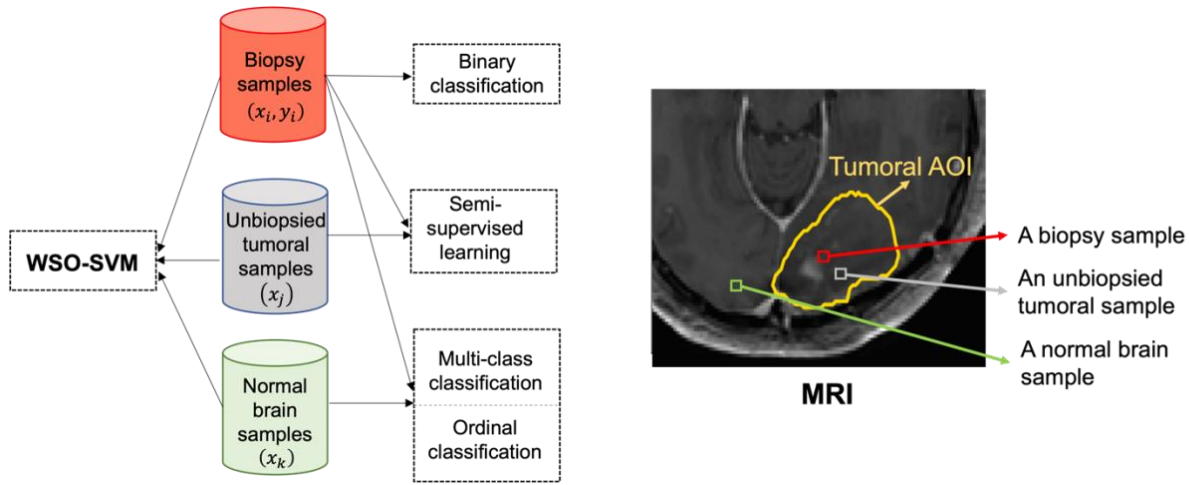

**Fig. 1 Different data sources that can be leveraged by WSO-SVM and existing ML algorithms.**

WSO-SVM is a novel ordinal classifier based on SVM. Unlike the existing algorithms that only utilize labeled samples from each class (e.g., normal brain samples—class 0, and biopsy samples—class 1 & 2), WSO-SVM introduces a unique optimization formulation to allow the incorporation of unlabeled tumoral samples (class 1 or 2, not 0). This helps identify accurate classification boundaries and improve prediction performance. The development of WSO-SVM is significant as it represents the first method capable of integrating multiple data sources, including biopsy samples, unlabeled tumoral samples, and normal brain samples, to train a robust classifier for predicting regional genetic status using MRI. In our case study, we demonstrate the superior performance of WSO-SVM compared to a variety of ML algorithms. The clinical utility of this work lies in the non-invasive quantification of intra-tumoral genetic heterogeneity using MRI for individual patients. WSO-SVM enables the generation of regional prediction maps for GBM driver genes such as EGFR, PDGFRA, and PTEN across the entire tumoral AOI for each patient. These maps have practical implications in guiding therapy selection and predicting response to targeted therapies, such as EGFR inhibitors [7]. Furthermore, the predictive maps reveal the co-existence of genomically distinct tumor subpopulations within individual tumors, which can enhance our understanding and develop new approaches, such as adaptive therapy, to leverage the interplay and competition between different molecular subpopulations for therapeutic benefit [18].

## 2. Method

Fig 2 shows a pipeline of the proposed method whose components are discussed in subsequent sections.

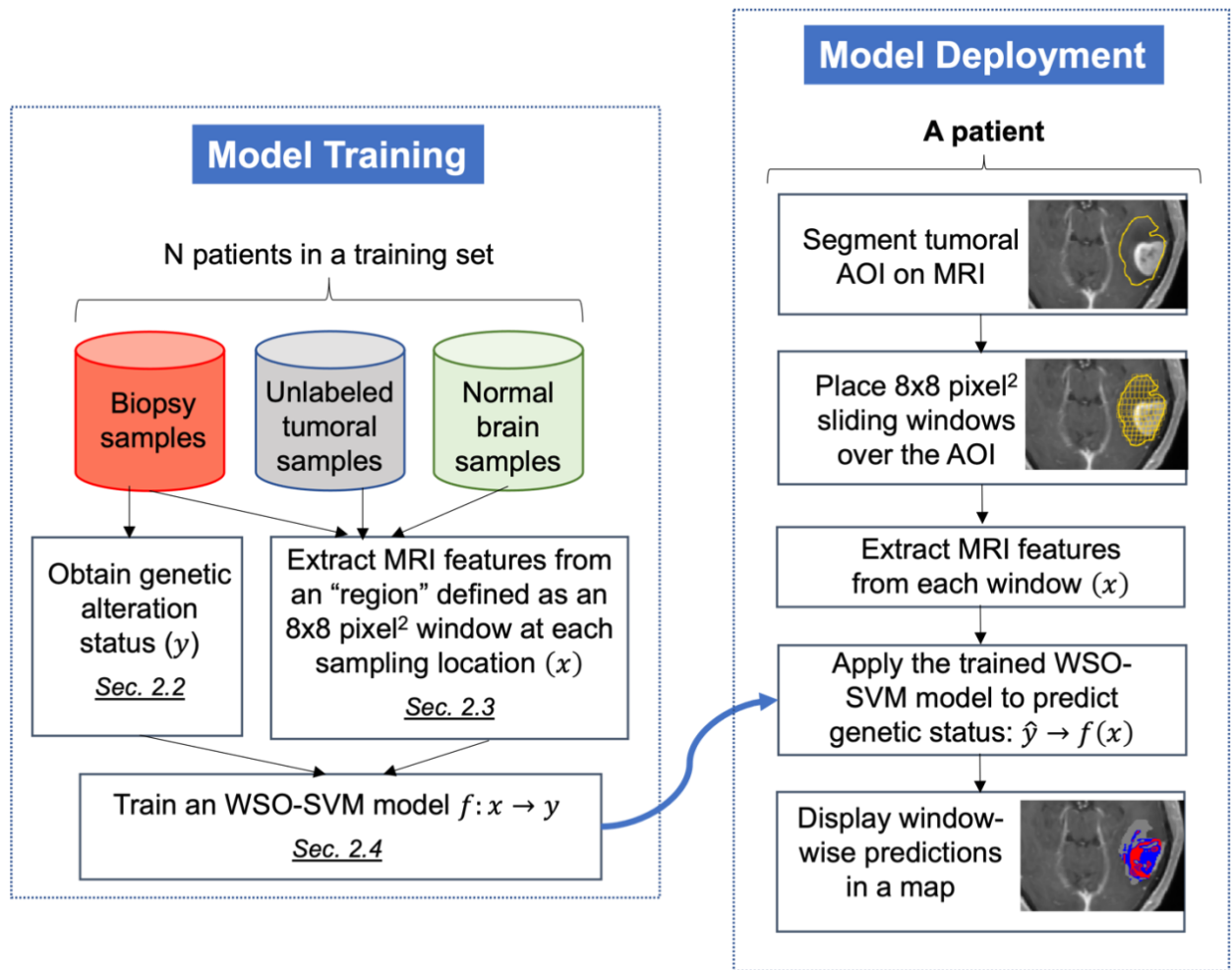

**Fig. 2** Pipeline of the proposed method. Left: model training; Right: model deployment.

## 2.1 Data collection

This study used data from a cohort of 74 GBM patients with IRB approval from Barrow Neurological Institute (BNI) and Mayo Clinic Arizona (MCA). These patients were prospectively recruited for the study. The recruitment period is from February 29, 2012, until present. All patients provided written informed consent. The data were accessed for research purposes from February 29, 2012, until present. A total of 318 biopsy samples were acquired from these patients (average: 4; range: 1-13). Each patient went through a pre-operative multiparametric MRI exam, from which five contrast images were obtained: T1-weighted contrast-enhanced image (T1+C), T2-weighted image (T2), mean diffusivity (MD), fractional anisotropy (FA), and relative cerebral blood volume (rCBV).

## 2.2. Biopsy sample analysis

Array CGH data was obtained for a subset of biopsy samples [19]. Whole exome sequencing (WES) was performed remaining biopsies and paired blood samples. Quality control was performed using the MultiQC toolkit. The aligned paired-end clean reads were processed using Burrows-Wheeler Aligner2 and GATK3 to remove low-quality reads and realign around indels. Somatic SNVs and indels were detected using a combination of six variant calling algorithms: Freebayes5, MuTect26, TNhaplotyper7, TNscope7, TNsnv7, and VarScan28. Somatic copy number and tumor purity were estimated from WES data using PureCN12. GISTIC213 analysis was performed to identify recurrently amplified or deleted genomic regions by integrating the results from individual patients.

We focused on three GBM driver genes: EGFR, PDGFRA, and PTEN. For each gene, we considered the gene is altered (class 2) if it has an abnormal CNV or is mutated, and non-altered (class 1) otherwise. For EGFR and PDGFRA, we followed the literature [19] and considered amplification as abnormal CNV; for PTEN, deletion or loss was considered as abnormal CNV [20]. To maximize the sample size in ML training, we included all available samples for each gene. There are 130/171, 53/238, and 206/109 biopsy samples with altered/non-altered EGFR, PDGFRA, and PTEN, respectively.

## 2.3 MRI preprocessing and feature extraction

Detailed MRI protocols and preprocessing approaches can be found in S1 Appendix. The same approaches have been used in our prior publications [2][7][21], which have shown robust performance.

The MRI features corresponding to each biopsy sample were extracted from a defined "region", i.e., an 8x8 pixel$^2$ window centered at the sampling location. This specific window size was thoughtfully chosen due to its approximate equivalence to the physical size of biopsy samples, ensuring an alignment between the MRI features and the genetic status derived from the biopsy. Moreover, prior research findings have supported the suitability of this window size for effectively capturing the intra-tumoral heterogeneity of GBM [2][7][19].

From this window, we extracted 280 features from five aforementioned MRI contrast images, which included statistical features and texture features using two well-established texture analysis algorithms, Gray-Level Co-occurrence Matrix (GLCM) [22] and Gabor Filters (GF) [23]. Please find names of these features in S1 Appendix. Fig 3 depicts the biological connection between genetic alterations and these imaging-phenotypic features. These features have been widely used in the radiomics literature for GBM to aid in diagnosis, prognosis, and prediction of genetics-related tumor characteristics, such as genetic subtypes and copy number variations [7][19][24–27].

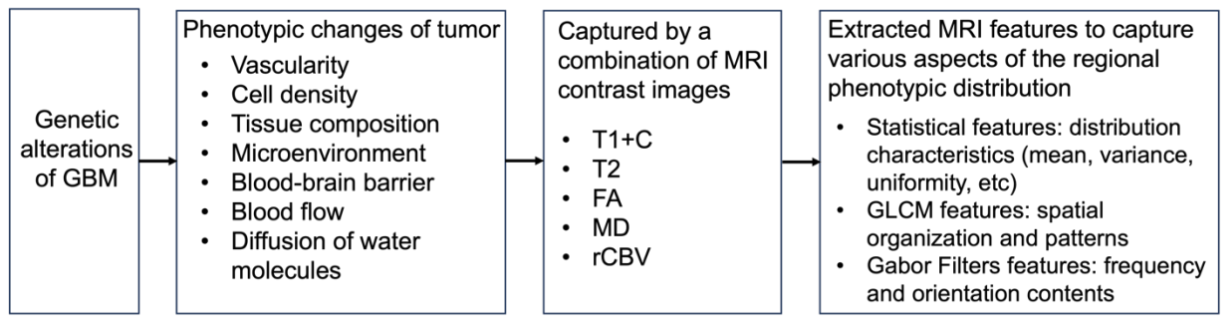

**Fig. 3 Biological connection between genetic alterations and imaging-phenotypic features.**

As shown in Fig 2, the training of WSO-SVM requires not only biopsy samples, but also unlabeled tumoral samples and normal brain samples. These are sampled from a pre-segmented tumoral AOI and the contralateral AOI based on the MRI of each patient, respectively. The tumoral AOI was segmented by following standard procedures [2][19], which is the union of the contrast-enhancing portion (CE) and the non-enhancing portion (NE) of the tumor. The contralateral AOI is located on the opposite side of the brain from the tumor and is considered "normal". To extract MRI features for these samples, the same approach as that used for biopsy samples was adopted.

The selection of unlabeled tumoral samples and normal brain samples was based on multi-fold considerations: (a) Representation of tumoral heterogeneity: Biologically, a GBM tumor includes a contrast-enhancing portion (CE) and a non-enhancing portion (NE). The former harbors proliferative tumor cells, while the latter harbors invading tumor to the surrounding brain tissue [28]. To ensure our unlabeled samples capture this biological heterogeneity of each tumor, an equal number of samples were taken from CE and NE. (b) Avoidance of outlier samples: We were careful to avoid selecting samples from areas that could be considered outliers. Notably, we excluded regions like necrosis, where the tissue characteristics significantly differ [28]. Additionally, for tumors located near fixed brain structures like the skull or cerebrospinal fluid, precautions were taken to prevent sample overlap with these structures. (c) Model accuracy and efficiency: Since unlabeled tumoral samples and normal brain samples are "auxiliary" samples to biopsies, their size should not be excessively larger even though acquiring these samples is much easier than biopsies. This is to prevent sample imbalance and potential dilution of the predominant influence of biopsy samples on model training. Therefore, we kept an equal number of unlabeled tumoral samples and normal brain samples, with their combined total aligning with that of biopsy samples. This choice also ensures the computational efficiency of model training.

Moreover, as depicted in Fig 2, when the trained WSO-SVM is applied to a patient, the goal is to generate a regional prediction map of the genetic status within the tumoral AOI. To accomplish this, an 8×8 pixel$^2$ sliding window with a stride size of one pixel was placed at each pixel within the tumoral AOI, and MRI features were extracted from each window.

## 2.4 Proposed WSO-SVM model

Let $D$ denote a training set that consists of $N$ patients. Assume there are $n_1$ and $n_2$ total biopsy samples from these patients with a gene of interest being non-altered ($y = 1$) and altered ($y = 2$), respectively. Let $x_i^{(1)}$ and $x_{i'}^{(2)}$ denote the MRI feature vectors for a biopsy sample in class 1 and 2, respectively; $i = 1, \dots, n_1; i' = 1, \dots, n_2$. Also, assume there are $m_{12}$ unlabeled tumoral samples ($y = 1$ or 2). Let $x_j^{(12)}$ denote the MRI feature vector for an unlabeled sample, $j = 1, \dots, m_{12}$. Additionally, assume there are $m_0$ normal brain samples ($y = 0$). Let $x_k^{(0)}$ denote the MRI feature vector for a normal brain sample, $k = 1, \dots, m_0$.

As illustrated in Fig 4, WSO-SVM maps the MRI feature vector of each sample, $x$, into a high-dimensional Reproducing Kernel Hilbert Space, $\phi(x)$, where a linear classifier $w^T\phi(x)$ is constructed to separate the three classes ($y = 0, 1, 2$) with largest possible margin, $2/\|w\|$, while also minimizing the empirical errors of samples that cannot be classified correctly, such as $\xi_i^{(1)}$, $\xi_{i'}^{(2)}, \zeta_j^{(12)}, \zeta_k^{(0)}$. The goal of training WSO-SVM to find the weight vector $w$ and two classification boundaries, $b_0$ and $b_1$. Formally, we construct WSO-SVM as the following optimization:

$$\min_{w,b_0,b_1,\xi,\zeta} \frac{1}{2}\|w\|$$

Subject to: $w^T\phi\left(x_i^{(1)}\right) - b_1 \leq -1 + \xi_i^{(1)}, \xi_i^{(1)} \geq 0, for\ i = 1, \ldots, n_1;$ (1)

$$w^T\phi\left(x_{i'}^{(2)}\right) - b_1 \geq 1 - \xi_{i'}^{(2)}, \xi_{i'}^{(2)} \geq 0, for\ i' = 1, \ldots, n_2;$$ (2)

$$\sum_{i=1}^{n_1} \xi_i^{(1)} + \sum_{i=1}^{n_2} \xi_{i'}^{(2)} \leq \epsilon;$$ (3)

$$w^T\phi\left(x_k^{(0)}\right) - b_0 \leq -1 + \zeta_k^{(0)}, \zeta_k^{(0)} \geq 0, k = 1, \ldots, m_0;$$ (4)

$$w^T\phi\left(x_j^{(12)}\right) - b_0 \geq 1 - \zeta_j^{(12)}, \zeta_j^{(12)} \geq 0, j = 1, \ldots, m'_{12}, m'_{12} = n_1 + n_2 + m_{12};$$ (5)

$$\sum_{k=1}^{m_0} \zeta_k^{(0)} + \sum_{j=1}^{m'_{12}} \zeta_j^{(12)} \leq e;$$ (6)

$$b_0 \leq b_1.$$ (7)

The objective function seeks to maximize the margin that separates different classes. The constraints (1)-(2) are designed to classify biopsy samples into classes 1 and 2, while introducing slack values, $\xi_i^{(1)}$, $\xi_{i'}^{(2)}$, to allow for some misclassification errors, which are bounded in (3). The constraints (4)-(5) are designed to classify normal brain samples into class 0, and to prevent unlabeled tumoral samples and biopsy samples from being classified as class 0, while introducing slack values, $\zeta_k^{(0)}$, $\zeta_j^{(12)}$, to allow for some misclassification errors, which are bounded in (6). The constraint in (7) is intended to retain the intrinsic order of the ordinal classes, 0, 1, and 2.

It is important to note that WSO-SVM is different from ordinal SVM in its ability to incorporate unlabeled tumoral samples. This is achieved by introducing a constraint in Eq. (5) to prevent the classification of these samples as normal brain samples (class 0). The inclusion of unlabeled tumoral samples helps better identify the classification boundary $b_0$, and also contributes to the estimation of the weight vector $w$, indirectly aiding in a better identification of $b_1$.

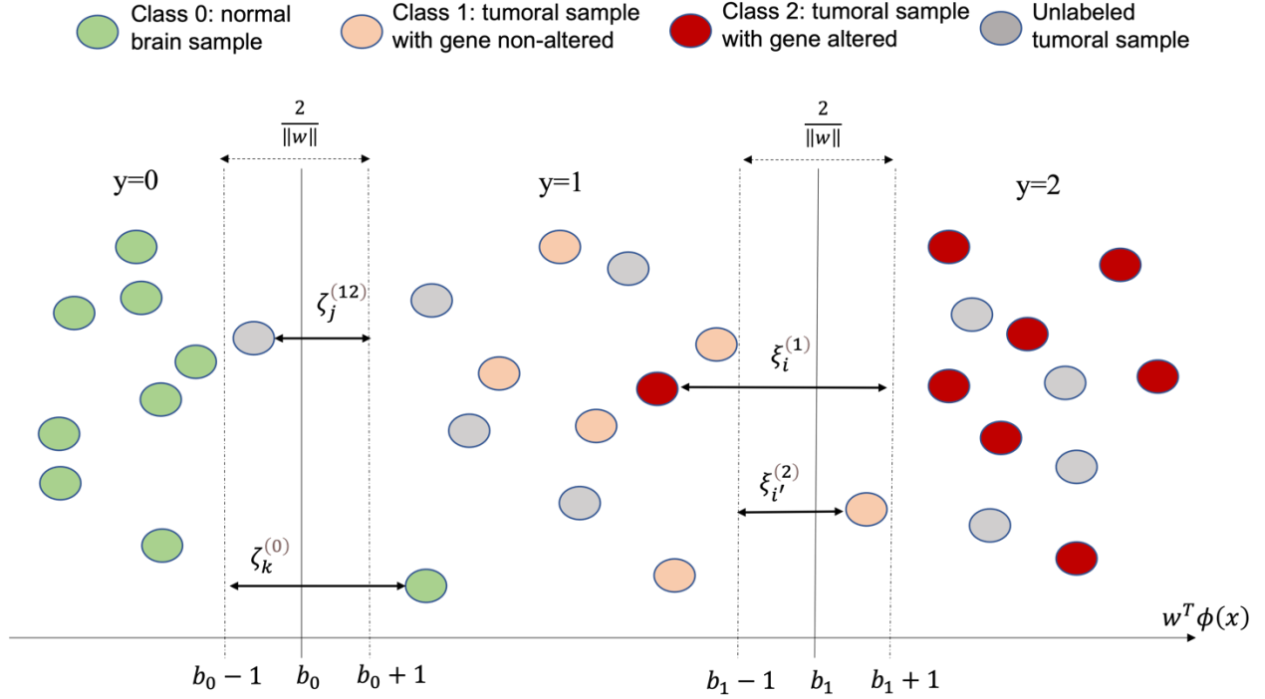

**Fig. 4: A graphical illustration of the model formulation of WSO-SVM**

It is easier to solve the WSO-SVM optimization in its dual form which is given in Proposition 1.

**Proposition 1:** The dual form of the primal WSO-SVM optimization problem in Eq. (1)-(7) is:

$$\min_{\alpha,\beta} \frac{1}{2}\gamma^T Y K Y \gamma - \sum_{i=1}^{n_1} \alpha_i^{(1)} - \sum_{i'=1}^{n_2} \alpha_{i'}^{(2)} - \sum_{k=1}^{m_0} \beta_k^{(0)} - \sum_{j=1}^{m'_{12}} \beta_j^{(12)},$$

subject to:

$$-\sum_{i=1}^{n_1} \alpha_i^{(1)} + \sum_{i'=1}^{n_2} \alpha_{i'}^{(2)} - \sum_{k=1}^{m_0} \beta_k^{(0)} + \sum_{j=1}^{m'_{12}} \beta_j^{(12)} = 0,$$

$$-\sum_{i=1}^{n_1} \alpha_i^{(1)} + \sum_{i'=1}^{n_2} \alpha_{i'}^{(2)} \geq 0,$$

$$0 \leq \alpha_i^{(1)} \leq C_1, i = 1, \ldots, n_1; 0 \leq \alpha_{i'}^{(2)} \leq C_1, i' = 1, \ldots, n_2,$$

$$0 \leq \beta_k^{(0)} \leq C_2, k = 1, \ldots, m_0; 0 \leq \beta_j^{(12)} \leq C_2, j = 1, \ldots, m'_{12},$$

where $\gamma = (\alpha_1^{(1)}, \ldots, \alpha_{n_1}^{(1)}, \alpha_1^{(2)}, \ldots, \alpha_{n_2}^{(2)}, \beta_1^{(0)}, \ldots, \beta_{m_0}^{(0)}, \beta_1^{(12)}, \ldots, \beta_{m_{12}'}^{(12)})$,

$Y = diag\left(\underbrace{-1,\ldots,-1}_{n_1}, \underbrace{1,\ldots,1}_{n_2}, \underbrace{-1,\ldots,-1}_{m_0}, \underbrace{1,\ldots,1}_{m_{12}'}\right)$, and $K$ is a covariance matrix with $K_{ij} = \phi(x_i)^T \phi(x_j) = k(x_i, x_j)$ that can be computed by a kernel function defined on the feature space. $C_1$ and $C_2$ are tuning parameters. (Proof in S1 Appendix.)

The dual problem is a convex quadratic programming problem, which can be solved by a standard quadratic optimization solver such as CPLEX.

Once the optimal solutions of $\alpha$ and $\beta$ in the dual problem are obtained, we can obtain the optimal coefficients in the primal problem, $w$, and further get $h(x) = -\sum_{i=1}^{n_1} \alpha_i^{(1)} k\left(x, x_i^{(1)}\right) + \sum_{i'=1}^{n_2} \alpha_{i'}^{(2)} k\left(x, x_{i'}^{(2)}\right) - \sum_{k=1}^{m_0} \beta_k^{(0)} k\left(x, x_k^{(0)}\right) + \sum_{j=1}^{m_{12}'} \beta_j^{(12)} k\left(x, x_j^{(12)}\right)$. Also, $b_0$ and $b_1$ can be estimated as: $b_0 = h(x) - y$ for any $(x, y)$ belonging to normal brain samples (or biopsy and unlabeled tumoral samples) whose corresponding $\beta^{(0)}$ (or $\beta^{(12)}$) satisfies $0 \leq \beta^{(0)}$ (or $\beta^{(12)}) \leq C_2$; $b_1 = h(x) - y$ for any $(x, y)$ belonging to non-altered biopsy samples (or altered biopsy samples) whose corresponding $\alpha^{(1)}$ (or $\alpha^{(2)}$) satisfies $0 \leq \alpha^{(1)}$ (or $\alpha^{(2)}) \leq C_1$. Then, we can obtain the discriminant functions for any new sample $x^*$, i.e., $f_0(x^*) = sign(h(x^*) - b_0)$ and $f_1(x^*) = sign(h(x^*) - b_1)$. The decision rule for classifying the new sample $x^*$ is: it belongs to class 2 if $f_1(x^*) \geq 0$, to class 1 if $f_1(x^*) < 0$ & $f_0(x^*) \geq 0$, and to class 0 if $f_0(x^*) < 0$.

<u>Training and cross validation (CV).</u> We used 10-fold CV to mitigate the risk of overfitting. To further reduce potential bias in evaluating model performance due to the specific fold division in CV, we repeated the CV procedure 30 times. We reported the model's average performance and the standard deviation across the 30 repetitions with the latter capturing uncertainty. Specifically, the biopsy samples were divided into 10 folds. In each iteration, WSO-SVM was trained based on 9 folds of the biopsy samples and randomly selected unlabeled tumoral samples and normal brain samples of the same size according to the considerations illustrated in Sec. 2.3.

Choice of tuning parameters: There are two key tuning parameters for WSO-SVM according to Proposition 1, $C_1$ and $C_2$. $C_1$ affects the classification boundary between biopsy samples in class 1 (gene not altered) and class 2 (gene altered). $C_2$ affects classification boundary between class 1 or 2 (comprising tumoral samples, both unlabeled and labeled) and class 0 (normal brain samples). Our experiments found that distinguishing between class 1/2 and class 0 was relatively easy, which also aligned with the intuition that discerning tumoral samples from normal brain samples should inherently be a formidable task. Therefore, we tuned $C_2$ on a coarser grid within the range of 0.01 to 100 and kept multiple settings that yielded >80% accuracy in differentiating class 1/2 from class 0. At each setting, we tuned $C_1$ on a finer grid between 0.01 and 100, and selected $C_1$ with the highest accuracy to differentiate class 1 and 2.

Generation of a regional predictive map of genetic status for each patient. To personalize the model toward each patient's data, we re-trained WSO-SVM under the previously found optimal tuning parameter setting but using randomly selected unlabeled tumoral samples and normal brain samples from the specific patient. Next, we applied the model to predict the gene status for each sliding window within the tumoral AOI of the patient, based on MRI features extracted from that window. The resulting predictions formed the predictive map for that patient.

Time complexity in training and deployment. As WSO-SVM adopted SVM as its base model, its time complexity in model training is similar to that of SVM [29], which ranges between $O(n^2 \times d)$ and $O(n^3 \times d)$, where $n$ is the sample size and $d$ is the feature dimension. Currently, we used quadratic programming to solve the WSO-SVM optimization, which can be further expedited by using more advanced optimization algorithms such as sequential Minimal optimization [30] and stochastic gradient descent [31]. While SVM-type of models are not the most computationally efficient, the training time complexity is acceptable and the performance gain over more efficient methods has made it an appealing choice for large datasets in various applications. In our application, the model training is done offline, which makes it feasible to train WSO-SVM on large datasets. During deployment, the trained model generates regional genetic characteristics within the tumoral area on a patient-by-patient basis. The time required to produce the prediction map for an individual patient is less than 30 seconds when executed on a standard desktop computer. This level of efficiency aligns well with the clinical use case, ensuring that the model can be deployed in a timely and practical manner.

## 2.5 Model interpretation

It is important to understand the contribution of different MRI features to the prediction made by WSO-SVM. While WSO-SVM can use either a linear or non-linear kernel, we found that a non-linear kernel produced better performance. Also, previous studies have shown that the relationship between MRI features and genetic status is highly non-linear [32]. To interpret the non-linear WSO-SVM, we utilized a popular, model-agnostic method called SHapley Additive exPlanations (SHAP) [33]. Essentially, SHAP estimates the contribution of a feature, referred to as the SHAP value, by computing the difference in the model's prediction when the feature is present versus absent. The higher the absolute SHAP value of a feature, the greater its impact on the prediction. In our study, we were more interested in the contribution of each MRI contrast image rather than individual features. Thus, we aggregated the feature-wise SHAP values to the contrast level.

## 2.6 Competing methods

We compared the performance of WSO-SVM with existing algorithms in several categories (using the same CV process):

- Binary classifiers: SVM, random forest (RF).
- Semi-supervised learning algorithms: transductive SVM (TSVM) [34], Laplacian SVM (LapSVM) [35], co-training [36], semi-supervised RF (semi-RF) [37].
- Multi-class classifiers: SVM, RF.
- Ordinal classifiers: ordinal SVM, ordinal RF
- Multi-task learning (MTL): regularized MTL (regMTL) [38], MTL Gaussian Process (MTL-GP) [39], MTL RF (MTL-RF) [40]. These are multi-class classification algorithms by coupling the models of the three GBM driver genes together.

# 3. Results

Table 1-3 summarize the average CV performance and standard deviation over 30 repeated experiments for each gene. Fig 5 compares WSO against the competing algorithm with the best accuracy in each category. WSO-SVM achieved the highest accuracy, sensitivity, and specificity for EGFR and PTEN. For PDGFRA, WSO-SVM achieved the highest accuracy and sensitivity, while its specificity is second highest after MTL-RF. However, the sensitivity of MTL-RE is very low (only 0.5). Due to the heavy class imbalance for PDGFRA, most existing algorithms struggle to achieve a reasonable sensitivity, whereas WSO-SVM did not have this issue. Among all the competing algorithms, random forest types of methods performed better in most cases. Moreover, the standard deviation of WSO-SVM is among the smallest over all the methods being compared. The magnitude of the standard deviation is also small, indicating that the model performance is quite stable (i.e., less uncertainty).

To assess the statistical significance of the performance gain for WSO-SVM, we performed a one-sided Wilcoxon rank-sum test to compare WSO-SVM against the competing algorithm with the overall best accuracy. For EGFR, WSO-SVM significantly outperformed multi-class RF in accuracy, sensitivity, and specificity ($p<0.001$, $p<0.001$, $p=0.002$). For PTEN, WSO-SVM significantly outperformed binary RF in accuracy, sensitivity, and specificity ($p<0.001$, $p<0.001$, $p<0.001$). For PDGFRA, WSO-SVM had significantly higher accuracy and sensitivity than MTL-RF ($p=0.04$, $p<0.001$), but its specificity was not significantly higher.

**Table 1: Classification performance of EGFR using CV based on biopsy samples**

| Category | Model | Accuracy | Sensitivity | Specificity |
|---|---|---|---|---|
| Binary classification | SVM | 0.69 (0.017) | 0.68 (0.026) | 0.70 (0.019) |
|  | RF* | 0.72 (0.012) | 0.73 (0.020) | 0.72 (0.018) |
| Semi-supervised learning | TSVM | 0.60 (0.025) | 0.61 (0.041) | 0.59 (0.030) |
|  | LapSVM | 0.68 (0.018) | 0.64 (0.025) | 0.70 (0.023) |
|  | Co-training* | 0.69 (0.019) | 0.73 (0.030) | 0.66 (0.025) |
|  | Semi-RF* | 0.69 (0.024) | 0.70 (0.031) | 0.68 (0.034) |
| Multi-class classification | SVM | 0.69 (0.015) | 0.56 (0.032) | 0.79 (0.020) |
|  | RF** | 0.74 (0.013) | 0.66 (0.023) | 0.79 (0.019) |
| Ordinal classification | Ordinal SVM* | 0.71 (0.012) | 0.71 (0.018) | 0.70 (0.018) |
|  | Ordinal RF | 0.64 (0.027) | 0.55 (0.032) | 0.70 (0.040) |
| Multi-task learning | RegMTL | 0.64 (0.015) | 0.64 (0.032) | 0.64 (0.029) |
|  | MTL-GP | 0.68 (0.032) | 0.67 (0.041) | 0.68 (0.034) |
|  | MTL-RF* | 0.70 (0.012) | 0.62 (0.015) | 0.77 (0.017) |
| **WSO-SVM** |  | **0.80 (0.013)** | **0.79 (0.020)** | **0.81 (0.015)** |

\* Best competing algorithm in each category     \*\* Overall best competing algorithm

WSO-SVM performed significantly better than the overall best competing algorithm in accuracy (p<0.001), sensitivity (p<0.001), and specificity (p=0.002) using a Wilcoxon rank-sum test.

**Table 2: Classification performance of PDGFRA using CV based on biopsy samples**

| Category | Model | Accuracy | Sensitivity | Specificity |
|---|---|---|---|---|
| Binary classification | SVM | 0.60 (0.028) | 0.61 (0.052) | 0.59 (0.033) |
|  | RF* | 0.64 (0.028) | 0.65 (0.067) | 0.64 (0.028) |
| Semi-supervised learning | TSVM | 0.54 (0.030) | 0.57 (0.095) | 0.53 (0.029) |
|  | LapSVM | 0.58 (0.031) | 0.64 (0.080) | 0.57 (0.029) |
|  | Co-training* | 0.62 (0.028) | 0.62 (0.055) | 0.62 (0.035) |
|  | Semi-RF | 0.61 (0.034) | 0.59 (0.063) | 0.61 (0.038) |
| Multi-class classification | SVM* | 0.65 (0.023) | 0.63 (0.054) | 0.65 (0.028) |
|  | RF* | 0.65 (0.022) | 0.60 (0.059) | 0.66 (0.027) |
| Ordinal classification | Ordinal SVM* | 0.63 (0.022) | 0.64 (0.047) | 0.63 (0.025) |
|  | Ordinal RF | 0.61 (0.036) | 0.51 (0.066) | 0.64 (0.048) |
| Multi-task learning | RegMTL | 0.52 (0.017) | 0.65 (0.066) | 0.49 (0.019) |
|  | MTL-GP | 0.59 (0.024) | 0.69 (0.046) | 0.57 (0.029) |
|  | MTL-RF** | 0.70 (0.019) | 0.50 (0.047) | 0.75 (0.024) |
| **WSO-SVM** |  | **0.71 (0.019)** | **0.70 (0.060)** | **0.72 (0.025)** |

\* Best competing algorithm in each category     \*\* Overall best competing algorithm

WSO-SVM performed significantly better than the overall best competing algorithm in accuracy (p=0.04) and sensitivity (p<0.001) using a Wilcoxon rank-sum test.

**Table 3: Classification performance of PTEN using CV based on biopsy samples**

| Category | Model | Accuracy | Sensitivity | Specificity |
|---|---|---|---|---|
| Binary classification | SVM | 0.40 (0.021) | 0.38 (0.032) | 0.42 (0.042) |
| | RF** | 0.71 (0.018) | 0.76 (0.026) | 0.64 (0.034) |
| Semi-supervised learning | TSVM | 0.57 (0.025) | 0.56 (0.033) | 0.58 (0.053) |
| | LapSVM | 0.58 (0.024) | 0.56 (0.032) | 0.60 (0.034) |
| | Co-training | 0.62 (0.026) | 0.62 (0.036) | 0.62 (0.043) |
| | Semi-RF* | 0.63 (0.022) | 0.65 (0.034) | 0.59 (0.042) |
| Multi-class classification | SVM | 0.60 (0.029) | 0.57 (0.033) | 0.66 (0.050) |
| | RF* | 0.67 (0.031) | 0.65 (0.032) | 0.70 (0.043) |
| Ordinal classification | Ordinal SVM* | 0.62 (0.027) | 0.63 (0.032) | 0.61 (0.041) |
| | Ordinal RF | 0.58 (0.027) | 0.54 (0.041) | 0.68 (0.042) |
| Multi-task learning | RegMTL | 0.54 (0.021) | 0.47 (0.033) | 0.65 (0.035) |
| | MTL-GP | 0.61 (0.030) | 0.62 (0.041) | 0.61 (0.026) |
| | MTL-RF* | 0.63 (0.017) | 0.64 (0.025) | 0.61 (0.024) |
| **WSO-SVM** | | **0.80 (0.017)** | **0.78 (0.022)** | **0.83 (0.026)** |

\* Best competing algorithm in each category      \*\* Overall best competing algorithm

WSO-SVM performed significantly better than the overall best competing algorithm in accuracy ($p<0.001$), sensitivity ($p<0.001$), and specificity ($p<0.001$) using a Wilcoxon rank-sum test.

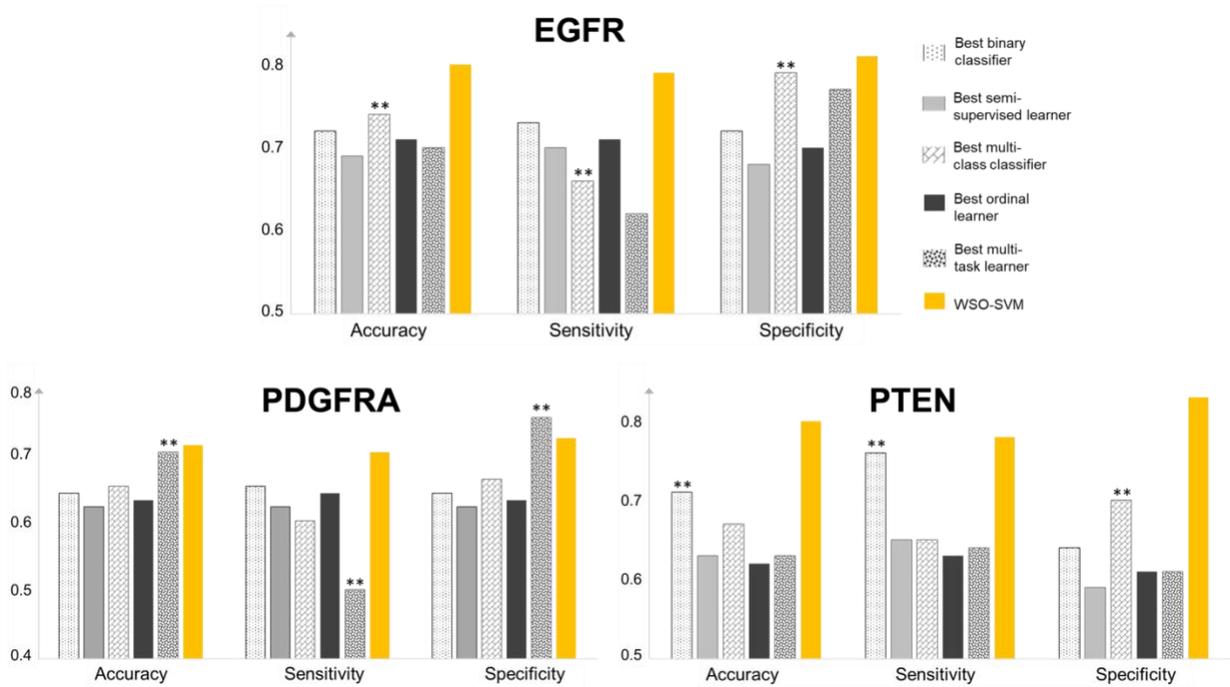

**Fig. 5 Classification performance of WSO-SVM in comparison with the best competing algorithm in each category. The overall best competing algorithm is highlighted by \*\*.**

Furthermore, Fig 6 shows the absolute SHAP values of the five MRI contrast images. It is evident that all contrast images contribute to the classification of each gene, but their relative contributions vary between genes. Further discussion will be provided in the next section.

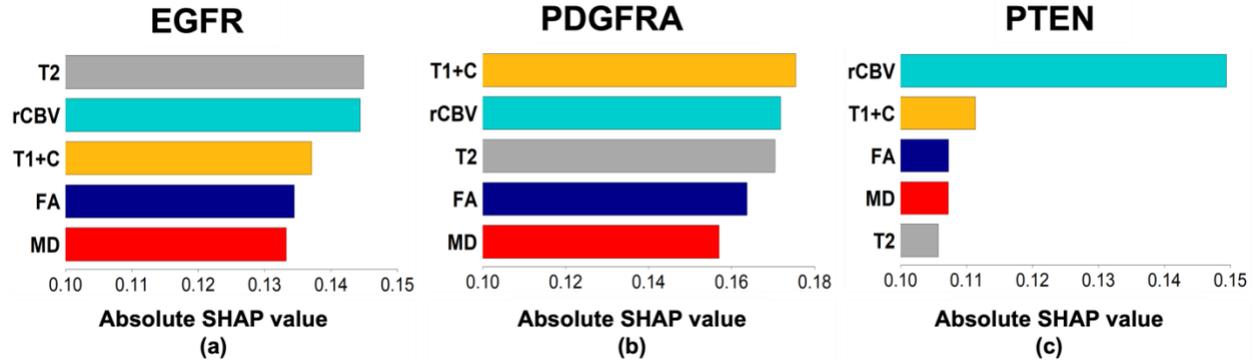

**Fig. 6 Contributions of MRI contrast images to the classification of (a) EGFR, (b) PDGFRA, and (c) PTEN, by WSO-SVM.**

Finally, the trained WSO-SVM models were used to generate prediction maps of the three genes for each patient. For demonstration, Fig 7 shows the prediction maps for four different patients. The alterations in EGFR and PDGFRA promote tumor growth. Thus, we showed their co-alteration patterns in one map. PTEN is a tumor suppressor gene, whose alteration is shown in a separate map. Patient A demonstrates predominant regions with EGFR alteration, with scattered regions of PDGFRA co-alteration; the PTEN map shows largely non-alteration. For patient B, the PTEN map shows an opposite pattern, whereas the EGFR & PDGFRA map demonstrates a similar pattern as patient A. In contrast to patient A and B, patient C demonstrates predominant regions with PDGFRA alteration. For patient D, the regions with EGFR & PDGFRA co-alteration are relatively concentrated compared to the other patients. These examples demonstrated the great extent of intra-tumoral genetic heterogeneity for each patient.

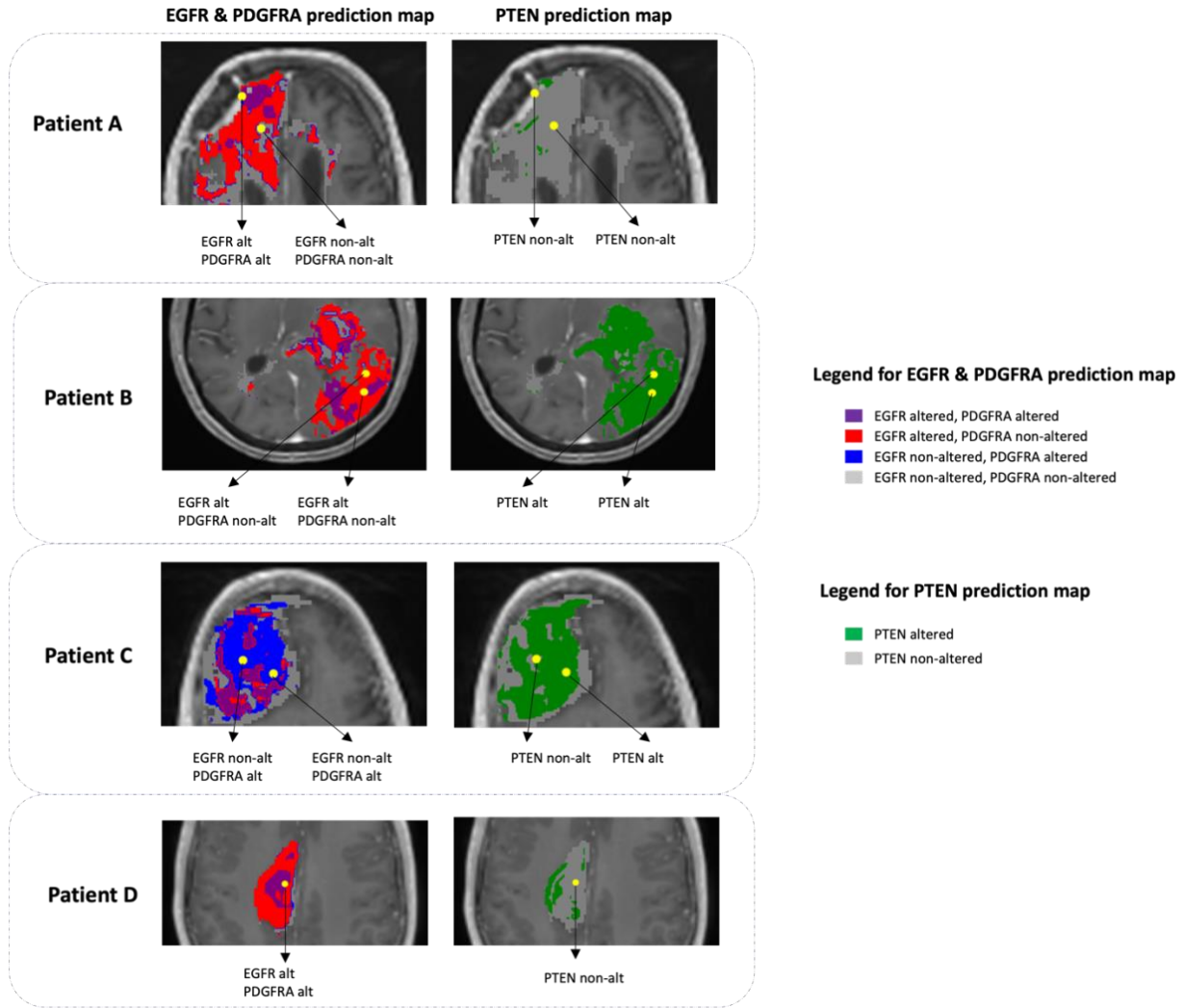

**Fig. 7: EGFR & PDGFRA prediction map (left column) and PTEN prediction map (right column) in tumoral AOI for four patients (rows). Yellow dots represent biopsy samples whose predicted gene statuses by WSO-SVM are reported underneath the maps (all predictions are correct).**

## 4. Discussion

Our results demonstrated that WSO-SVM surpasses a variety of existing ML algorithms for predicting the regional status of three GBM driver genes using MRI. To interpret WSO-SVM, the SHAP values in Fig 6 revealed the importance of each contrast in influencing WSO-SVM's prediction for each gene. Specifically, the model's predictions on EGFR were primarily influenced by T2 and rCBV, which aligns with prior research that found significant correlations of EGFR with T2 [19][41] and rCBV [11][19][42]. T1+C demonstrated the highest contribution to PDGFRA prediction. This is consistent with previous studies indicating that PDGFRA subpopulations tend to localize in CE with relatively less infiltration into NE, in comparison to EGFR [43]. For PTEN, the model's prediction received the greatest contribution from rCBV. Prior studies have highlighted the correlation between PTEN and rCBV, particularly when co-existing with EGFR alterations [44].

The prediction maps in Fig 7 and in S1 Fig for other patients in our dataset provided strong evidence of the extensive intra-tumoral genetic heterogeneity in each patient. While intra-tumoral genetic heterogeneity in GBM is well-documented in literature, practical methods for quantifying this heterogeneity are lacking. Biopsy samples, which can only be obtained from a few locations of the brain, leave many regions uncharacterized. This study introduces WSO-SVM as a non-invasive approach to predict regional genetic status across the entire tumoral AOI for each patient using MRI.

The clinical utility of the prediction maps for GBM driver genes, EGFR, PDGFRA, and PTEN, is multi-fold. First, these driver genes have been investigated as therapeutic targets for GBM. EGFR is one of the most commonly altered gene drivers in GBM and has been implicated in several pathogenic mechanisms. Targeted drug therapies, including those directed at EGFR and other receptor tyrosine kinases (RTKs) like PDGFRA, have been developed [11][12]. However, the clinical outcomes of current therapies are unsatisfactory for most patients due to the limited information obtained from sparse biopsy samples, which cannot fully capture the genetic landscape of each patient's tumor. With the capability provided by WSO-SVM, there is an opportunity to optimize therapy selection for each patient and provide better prognostic information regarding their response to treatment. This holds great potential for improving patient outcomes and tailoring therapies to individual genetic characteristics.

Moreover, this study goes beyond individual gene predictions and allows for the simultaneous prediction of multiple GBM driver genes. Interactions between tumor subpopulations within GBM tumors are increasingly acknowledged for their impact on biological behavior, therapeutic response, and local phenotypic expression. Although such interactions have been extensively studied in non-CNS tumors, their exploration in GBM remains limited. Existing studies have primarily focused on the heterogeneous expression of receptor tyrosine kinase (RTK) aberrations, such as EGFR and PDGFRA amplifications. For instance. Inda *et al*. [45] showed that a minority subpopulation expressing EGFR-vIII could potentiate a majority subpopulation expressing wild-type EGFR to enhance growth, survival, and drug resistance. Szerlip *et al.* [46] observed cooperation between subpopulations expressing EGFR or PDGFRA amplifications, requiring combined inhibition for pathway attenuation *in vitro*. Fiorenzo *et al*. [47] suggests that *in vivo* and human studies are needed to fully understand subpopulation interactions' impact on tumor growth. These interactions between subpopulations pose significant challenges for current treatment strategies and clinical trials that focus on single drug targets, such as EGFR [48]. By providing the capability to predict multiple GBM driver genes simultaneously, our study offers insights into these complex interactions and addresses the need for a more comprehensive understanding of tumor heterogeneity in GBM to develop future, advanced therapy [18,49].

This study has several limitations. First, the biopsy sample size is relatively small. This is due to the highly invasive nature of acquiring these samples from patients' brains. In the literature of integrating MRI and brain biopsy data for machine learning models, the typical sample size falls within the range of 82-244 [2][5][7][11]-[13]. While our study included 318 biopsies, a size comparably larger than these existing studies, it remains relatively modest when compared to domains where sample collection is more accessible. To alleviate this problem, the WSO-SVM model was designed to incorporate unlabeled tumoral samples and normal brain samples. However, further research is imperative to validate the generalizability of WSO-SVM on a more extensive and diverse population. A related issue is that our performance evaluation was based on CV. Using external datasets to further validate our model is highly necessary. There is currently no publicly available dataset with the same nature of our dataset, due to the invasive nature of biopsy acquisition and the time-consuming process of patient consent, surgical procedures, genetic analysis, and image preprocessing. Nevertheless, our team is currently collecting more data and preparing for subsequent validations of the model. This paper serves as a starting point in addressing a critical issue of non-invasive quantification for intra-tumoral genetic heterogeneity using MRI and a novel machine learning model WSO-SVM.

Second, it is important to acknowledge that while our study establishes correlations between genetic alterations and imaging-phenotypic features, it does not establish causal relationships. Experimental validation of causal relationships, which may involve creating specific genetic alterations in animal models and observing their effects on imaging phenotypes, remains a critical step to confirm and gain a deeper understanding of the underlying cancer mechanisms.

Third, while we have provided some discussions on the potentials of using the method developed in this paper to help therapeutic selection and develop advanced therapy to improve patient outcomes, this paper focused on the research phase of the method development. Clinical validation in real-world setting is necessary to establish the actual utility and benefit of the proposed method. Such validation could encompass clinical trials designed to compare patient outcomes, such as treatment response and survival, between cohorts undergoing standard clinical protocols for therapeutic selection and those benefitting from the additional guidance provided by the regional genetic prediction maps generated by our method.

Last but not least, the WSO-SVM model has several aspects for improvement. For instance, WSO-SVM can incorporate unlabeled tumoral samples and normal brain samples. Currently, these samples were selected based on considerations illustrated in Sec. 2.3. This selection method can be refined by integrating more advanced computational strategies that take uncertainty and diversity into account [50] and by considering patient demographic information [51]. Also, WSO-SVM relies on texture features extracted from MRI as input, which may be influenced by imaging quality. Uncertainty quantification of WSO-SVM predictions considering input uncertainty is important, and a Bayesian version of the model could address this issue. Also, developing robust predictive models that are insensitive to input uncertainty would have greater clinical utility.

# 5. Conclusion

We developed a data-inclusive WSO-SVM model to predict regional genetic alteration status within each GBM tumor using MRI. This study demonstrated the feasibility of using MRI and WSO-SVM to enable non-invasive prediction of regional genetic alteration for each patient, which can inform future adaptive therapies for individualized oncology.

# Acknowledgments

We are grateful to all of those who have contributed to elements of this work, particularly the many *surgeons* for collecting the biopsies, with special thanks to Kris Smith, Peter Nakaji, Bernard Bendok, Devi Patra, and Richard Zimmerman, Klimet Donev, the *glioma biopsy protocol teams* for aiding in the logistics ensuring integrity of the biopsy samples and screenshots and in aiding with clinical data abstraction, including Barrett Anderies, Jessica Bauer, Spencer Bayless, Hend Bcharach, Regina Becker, Sameer Channer, Brenden Doyle, Lysette Elsner, Lily Esaleh, Ashlyn Gonzales, Crystal Harris, Morgan Hatlestead, Ryan Hess, Sandra Johnston, Yvette Lassiter-Morris, Julia Lorence, Ashley Napier, Ashley Nespodzany, Sejal Shanbhag, Sarah Van Dijk, Scott Whitmire, and finally other past and current members of the *image analysis team*, special mention of Cassandra Rickertsen and Lisa Paulson for their leadership.

# Author Contributions

Conceptualization: LW, KRS, NLT, LSH, JL

Data curation: FD, LC, CPS, GDL, KWS, JU, AH, PRJ, CK, RSZ, DPP, BRB, KAS, PN, KD, LCB, MMM, MC, AI, NLT, LSH

Formal analysis: LW, FD, LC, CPS, GDL, KWS, JU, AH, PRJ

Funding Acquisition: KRS, NLT, LSH, JL

Methodology: LW, KRS, NLT, LSH, JL

Project Administration: KRS, NLT, LSH, JL

Supervision: KRS, NLT, LSH, JL

Validation: LW, HW, AI, KRS, NLT, LSH, JL

Visualization: LW, HW, JL

Writing – original draft: LW, AH, NLT, LSH, JL

Writing – review & editing: LW, HW, KRS, NLT, LSH, JL

# References

1. Stupp R, Mason WP, van den Bent MJ, Weller M, Fisher B, Taphoorn MJB, et al. Radiotherapy plus concomitant and adjuvant temozolomide for glioblastoma. New England journal of medicine. 2005;352: 987–996.

2. Hu LS, Ning S, Eschbacher JM, Gaw N, Dueck AC, Smith KA, et al. Multi-parametric MRI and texture analysis to visualize spatial histologic heterogeneity and tumor extent in glioblastoma. PLoS One. 2015;10: e0141506.

3. Hu LS, Hawkins-Daarud A, Wang L, Li J, Swanson KR. Imaging of intratumoral heterogeneity in high-grade glioma. Cancer Lett. 2020;477: 97–106.

4. Swanson KR, Bridge C, Murray JD, Alvord Jr EC. Virtual and real brain tumors: using mathematical modeling to quantify glioma growth and invasion. J Neurol Sci. 2003;216: 1–10.

5. Baldock AL, Ahn S, Rockne R, Johnston S, Neal M, Corwin D, et al. Patient-specific metrics of invasiveness reveal significant prognostic benefit of resection in a predictable subset of gliomas. PLoS One. 2014;9: e99057.

6. Marusyk A, Almendro V, Polyak K. Intra-tumour heterogeneity: a looking glass for cancer? Nat Rev Cancer. 2012;12: 323–334.

7. Hu LS, Wang L, Hawkins-Daarud A, Eschbacher JM, Singleton KW, Jackson PR, et al. Uncertainty quantification in the radiogenomics modeling of EGFR amplification in glioblastoma. Sci Rep. 2021;11: 1–14.

8. Kha Q-H, Le V-H, Hung TNK, Le NQK. Development and Validation of an Efficient MRI Radiomics Signature for Improving the Predictive Performance of 1p/19q Co-Deletion in Lower-Grade Gliomas. Cancers (Basel). 2021;13. doi:10.3390/cancers13215398

9. Lam LHT, Do DT, Diep DTN, Nguyet DLN, Truong QD, Tri TT, et al. Molecular subtype classification of low-grade gliomas using magnetic resonance imaging-based radiomics and machine learning. NMR Biomed. 2022;35: e4792. doi:10.1002/nbm.4792

10. Akbari H, Bakas S, Pisapia JM, Nasrallah MP, Rozycki M, Martinez-Lage M, et al. In vivo evaluation of EGFRvIII mutation in primary glioblastoma patients via complex multiparametric MRI signature. Neuro Oncol. 2018;20: 1068–1079.

11. Tykocinski ES, Grant RA, Kapoor GS, Krejza J, Bohman L-E, Gocke TA, et al. Use of magnetic perfusion-weighted imaging to determine epidermal growth factor receptor variant III expression in glioblastoma. Neuro Oncol. 2012;14: 613–623.

12. Kickingereder P, Bonekamp D, Nowosielski M, Kratz A, Sill M, Burth S, et al. Radiogenomics of glioblastoma: machine learning–based classification of molecular characteristics by using multiparametric and multiregional MR imaging features. Radiology. 2016;281: 907–918.

13. Chen H, Lin F, Zhang J, Lv X, Zhou J, Li Z-C, et al. Deep Learning Radiomics to Predict PTEN Mutation Status From Magnetic Resonance Imaging in Patients With Glioma. Front Oncol. 2021;11.

14. Zhu X, Goldberg AB. Introduction to Semi-Supervised Learning. Springer International Publishing; 2009. doi:10.1007/978-3-031-01548-9

15. Chu W, Keerthi SS. Support Vector Ordinal Regression. Neural Comput. 2007;19: 792–815. Available: http://direct.mit.edu/neco/article-pdf/19/3/792/816834/neco.2007.19.3.792.pdf

16. Chu W, Uk ZUA, Williams CKI. Gaussian Processes for Ordinal Regression Zoubin Ghahramani. Journal of Machine Learning Research. 2005.

17. Shashua A, Levin A. Ranking with Large Margin Principle: Two Approaches*.

18. Gatenby RA, Silva AS, Gillies RJ, Frieden BR. Adaptive therapy. Cancer Res. 2009;69: 4894–4903. doi:10.1158/0008-5472.CAN-08-3658

19. Hu LS, Ning S, Eschbacher JM, Baxter LC, Gaw N, Ranjbar S, et al. Radiogenomics to characterize regional genetic heterogeneity in glioblastoma. Neuro Oncol. 2017;19: 128–137.

20. Koul D. PTEN signaling pathways in glioblastoma. Cancer Biol Ther. 2008;7: 1321–1325.

21. Gaw N, Hawkins-Daarud A, Hu LS, Yoon H, Wang L, Xu Y, et al. Integration of machine learning and mechanistic models accurately predicts variation in cell density of glioblastoma using multiparametric MRI. Sci Rep. 2019;9: 10063. doi:10.1038/s41598-019-46296-4

22. Haralick RM, Shanmugam K, Dinstein IH. Textural features for image classification. IEEE Trans Syst Man Cybern. 1973; 610–621.

23. Feichtinger HG, Strohmer T. Gabor analysis and algorithms: Theory and applications. Springer Science & Business Media; 2012.

24. Lewis MA, Ganeshan B, Barnes A, Bisdas S, Jaunmuktane Z, Brandner S, et al. Filtration-histogram based magnetic resonance texture analysis (MRTA) for glioma IDH and 1p19q genotyping. Eur J Radiol. 2019;113: 116–123. doi:10.1016/j.ejrad.2019.02.014

25. Vamvakas A, Williams SC, Theodorou K, Kapsalaki E, Fountas K, Kappas C, et al. Imaging biomarker analysis of advanced multiparametric MRI for glioma grading. Phys Med. 2019;60: 188–198. doi:10.1016/j.ejmp.2019.03.014

26. Ryu YJ, Choi SH, Park SJ, Yun TJ, Kim J-H, Sohn C-H. Glioma: application of whole-tumor texture analysis of diffusion-weighted imaging for the evaluation of tumor heterogeneity. PLoS One. 2014;9: e108335. doi:10.1371/journal.pone.0108335

27. Alis D, Bagcilar O, Senli YD, Isler C, Yergin M, Kocer N, et al. The diagnostic value of quantitative texture analysis of conventional MRI sequences using artificial neural networks in grading gliomas. Clin Radiol. 2020;75: 351–357. doi:10.1016/j.crad.2019.12.008

28. Eidel O, Burth S, Neumann J-O, Kieslich PJ, Sahm F, Jungk C, et al. Tumor Infiltration in Enhancing and Non-Enhancing Parts of Glioblastoma: A Correlation with Histopathology. PLoS One. 2017;12: e0169292. doi:10.1371/journal.pone.0169292

29. Chapelle O. Training a support vector machine in the primal. Neural Comput. 2007;19: 1155–78. doi:10.1162/neco.2007.19.5.1155

30. Platt J. Sequential Minimal Optimization: A Fast Algorithm for Training Support Vector Machines. 1998 Apr. Available: https://www.microsoft.com/en-us/research/publication/sequential-minimal-optimization-a-fast-algorithm-for-training-support-vector-machines/

31. Bishop CM, Nasrabadi NM. Pattern recognition and machine learning. Springer; 2006.

32. Ahn SJ, Kwon H, Yang JJ, Park M, Cha YJ, Suh SH, et al. Contrast-enhanced T1-weighted image radiomics of brain metastases may predict EGFR mutation status in primary lung cancer. Sci Rep. 2020;10. doi:10.1038/s41598-020-65470-7

33. Lundberg SM, Lee S-I. A unified approach to interpreting model predictions. Proceedings of the 31st international conference on neural information processing systems. 2017. pp. 4768–4777.

34. Collobert R, Sinz F, Weston J, Bottou L, Joachims T. Large scale transductive SVMs. Journal of Machine Learning Research. 2006;7.

35. Belkin M, Niyogi P, Sindhwani V. Manifold regularization: A geometric framework for learning from labeled and unlabeled examples. Journal of machine learning research. 2006;7.

36. Zhou Y, Goldman S. Democratic co-learning. 16th IEEE International Conference on Tools with Artificial Intelligence. IEEE; 2004. pp. 594–602.

37. Leistner C, Saffari A, Santner J, Bischof H. Semi-Supervised Random Forests. IEEE 12th international conference on computer vision. 2009. pp. 506–513.

38. Cao H, Zhou J, Schwarz E. RMTL: an R library for multi-task learning. Bioinformatics. 2019;35: 1797–1798.

39. Williams C, Bonilla E v, Chai KM. Multi-task Gaussian process prediction. Adv Neural Inf Process Syst. 2007; 153–160.

40. Linusson H. Multi-output Random Forests. 2013.

41. Aghi M, Gaviani P, Henson JW, Batchelor TT, Louis DN, Barker FG. Magnetic resonance imaging characteristics predict epidermal growth factor receptor amplification status in glioblastoma. Clinical Cancer Research. 2005;11: 8600–8605.

42. Gupta A, Young RJ, Shah AD, Schweitzer AD, Graber JJ, Shi W, et al. Pretreatment dynamic susceptibility contrast MRI perfusion in glioblastoma: prediction of EGFR gene amplification. Clin Neuroradiol. 2015;25: 143–150.

43. Snuderl M, Fazlollahi L, Le LP, Nitta M, Zhelyazkova BH, Davidson CJ, et al. Mosaic amplification of multiple receptor tyrosine kinase genes in glioblastoma. Cancer Cell. 2011;20: 810–817.

44. Ryoo I, Choi SH, Kim J-H, Sohn C-H, Kim SC, Shin HS, et al. Cerebral blood volume calculated by dynamic susceptibility contrast-enhanced perfusion MR imaging: preliminary correlation study with glioblastoma genetic profiles. PLoS One. 2013;8: e71704.

45. Inda M-M, Bonavia R, Mukasa A, Narita Y, Sah DWY, Vandenberg S, et al. Tumor heterogeneity is an active process maintained by a mutant EGFR-induced cytokine circuit in glioblastoma. Genes Dev. 2010;24: 1731–1745. doi:10.1101/gad.1890510

46. Szerlip NJ, Pedraza A, Chakravarty D, Azim M, McGuire J, Fang Y, et al. Intratumoral heterogeneity of receptor tyrosine kinases EGFR and PDGFRA amplification in glioblastoma defines subpopulations with distinct growth factor response. Proc Natl Acad Sci U S A. 2012;109: 3041–6. doi:10.1073/pnas.1114033109

47. Fiorenzo P, Mongiardi MP, Dimitri D, Cozzolino M, Ferri A, Montano N, et al. HIF1-positive and HIF1-negative glioblastoma cells compete in vitro but cooperate in tumor growth in vivo. Int J Oncol. 2010;36: 785–791. doi:10.3892/ijo_00000554

48. Hegi ME, Rajakannu P, Weller M. Epidermal growth factor receptor: a re-emerging target in glioblastoma. Curr Opin Neurol. 2012;25: 774–779. doi:10.1097/WCO.0b013e328359b0bc

49. Bonavia R, Inda M-M, Cavenee WK, Furnari FB. Heterogeneity Maintenance in Glioblastoma: A Social Network. Cancer Res. 2011;71: 4055–4060. doi:10.1158/0008-5472.CAN-11-0153

50. Etikan I. Sampling and Sampling Methods. Biom Biostat Int J. 2017;5. doi:10.15406/bbij.2017.05.00149

51. Elfil M, Negida A. Sampling methods in Clinical Research; an Educational Review. Emerg (Tehran). 2017;5: e52.

# Supporting Information

**Proof of Proposition 1**

Let $\alpha_i^{(1)}, \alpha_i^{(2)}, \beta_j^{(12)}, \beta_k^{(3)}, A_i^{(1)}, A_i^{(2)}, B_j^{(12)}, B_k^{(3)}, \mu \geq 0$ be Lagrangian multipliers and $C_1$ and $C_2$ be tuning parameters. The Lagrangian for the primal WSO-SVM optimization in Eq. (1)-(7) is

$$L = \frac{1}{2}w^T w + \sum_{i=1}^{n_1} \alpha_i^{(1)} \left(w^T \phi\left(x_i^{(1)}\right) - b_1 + 1 - \xi_i^{(1)}\right) - \sum_{i'=1}^{n_2} \alpha_{i'}^{(2)} \left(w^T \phi\left(x_{i'}^{(2)}\right) - b_1 - 1 + \xi_{i'}^{(2)}\right)$$

$$+ \sum_{k=1}^{m_0} \beta_k^{(0)} \left(w^T \phi\left(x_k^{(0)}\right) - b_0 + 1 - \zeta_k^{(0)}\right) - \sum_{j=1}^{m'_{12}} \beta_j^{(12)} \left(w^T \phi\left(x_j^{(12)}\right) - b_0 - 1 + \zeta_j^{(12)}\right) +$$

$$C_1 \left(\sum_{i=1}^{n_1} \xi_i^{(1)} + \sum_{i'=1}^{n_2} \xi_{i'}^{(2)}\right) + C_2 \left(\sum_{k=1}^{m_0} \zeta_k^{(0)} + \sum_{j=1}^{m'_{12}} \zeta_j^{(12)}\right) - \sum_{i=1}^{n_1} A_i^{(1)} \xi_i^{(1)} - \sum_{i'=1}^{n_2} A_{i'}^{(2)} \xi_{i'}^{(2)} -$$

$$\sum_{k=1}^{m_0} B_k^{(0)} \zeta_k^{(0)} - \sum_{j=1}^{m'_{12}} B_j^{(12)} \zeta_j^{(12)} + \mu(b_0 - b_1). \quad (8)$$

Then the optimal solution of the primal problem in Eq. (1)-(7) is equivalent to the solution of the following optimization:

$$\max_{\alpha,\beta,A,B,\mu} \min_{w,b,\xi,\zeta} L. \quad (9)$$

The KKT conditions for the primal problem require the following to hold:

$$\nabla_w L = w + \sum_{i=1}^{n_1} \alpha_i^{(1)} \phi\left(x_i^{(1)}\right) - \sum_{i'=1}^{n_2} \alpha_{i'}^{(2)} \phi\left(x_{i'}^{(2)}\right) + \sum_{k=1}^{m_0} \beta_k^{(0)} \phi\left(x_k^{(0)}\right) - \sum_{j=1}^{m'_{12}} \beta_j^{(12)} \phi\left(x_j^{(12)}\right) = 0,$$

$$\nabla_{b_1} L = -\sum_{i=1}^{n_1} \alpha_i^{(1)} + \sum_{i'=1}^{n_2} \alpha_{i'}^{(2)} - \mu = 0,$$

$$\nabla_{b_0} L = -\sum_{k=1}^{m_0} \beta_k^{(0)} + \sum_{j=1}^{m'_{12}} \beta_j^{(12)} + \mu = 0,$$

$$\nabla_{\xi_i^{(1)}} L = -\alpha_i^{(1)} + C_1 - A_i^{(1)} = 0, \ i = 1, \ldots, n_1,$$

$$\nabla_{\xi_{i'}^{(2)}} L = -\alpha_{i'}^{(2)} + C_1 - A_{i'}^{(2)} = 0, \ i' = 1, \ldots, n_2,$$

$$\nabla_{\zeta_k^{(0)}} L = -\beta_k^{(0)} + C_2 - B_k^{(0)} = 0, k = 1,\ldots,m_0.$$

$$\nabla_{\zeta_j^{(12)}} L = -\beta_j^{(12)} + C_2 - B_j^{(12)} = 0, \ j = 1,\ldots,m'_{12},$$

Then we have

$$w = -\sum_{i=1}^{n_1} \alpha_i^{(1)} \phi\left(x_i^{(1)}\right) + \sum_{i'=1}^{n_2} \alpha_{i'}^{(2)} \phi\left(x_{i'}^{(2)}\right) - \sum_{k=1}^{m_0} \beta_k^{(0)} \phi\left(x_k^{(0)}\right) + \sum_{j=1}^{m'_{12}} \beta_j^{(12)} \phi\left(x_j^{(12)}\right), \quad (10)$$

$$\mu = -\sum_{i=1}^{n_1} \alpha_i^{(1)} + \sum_{i'=1}^{n_2} \alpha_{i'}^{(2)}, \quad (11)$$

$$\mu = \sum_{k=1}^{m_0} \beta_k^{(0)} - \sum_{j=1}^{m'_{12}} \beta_j^{(12)}, \quad (12)$$

$$A_i^{(1)} = -\alpha_i^{(1)} + C_1, \ i = 1,\ldots,n_1, \quad (13)$$

$$A_{i'}^{(2)} = -\alpha_{i'}^{(2)} + C_1, \ i' = 1,\ldots,n_2, \quad (14)$$

$$B_k^{(0)} = -\beta_k^{(0)} + C_2, \ k = 1,\ldots,m_0, \quad (15)$$

$$B_j^{(12)} = -\beta_j^{(12)} + C_2, \ j = 1,\ldots,m'_{12}. \quad (16)$$

Inserting Eq. (11)-(16) into the optimization in Eq. (9), after simplification we can get

$$\max_{\alpha,\beta} L = \frac{1}{2} w^T w + \sum_{i=1}^{n_1} \alpha_i^{(1)} \left(w^T \phi\left(x_i^{(1)}\right) + 1\right) - \sum_{i'=1}^{n_2} \alpha_{i'}^{(2)} \left(w^T \phi\left(x_{i'}^{(2)}\right) - 1\right) +$$

$$\sum_{k=1}^{m_0} \beta_k^{(0)} \left(w^T \phi\left(x_k^{(0)}\right) + 1\right) - \sum_{j=1}^{m'_{12}} \beta_j^{(12)} \left(w^T \phi\left(x_j^{(12)}\right) - 1\right). \quad (17)$$

Furthermore, inserting Eq. (10) into the optimization in Eq. (17), we can have

$$\max_{\alpha,\beta} L = -\frac{1}{2} \gamma^T Y K Y \gamma + \sum_{i=1}^{n_1} \alpha_i^{(1)} + \sum_{i'=1}^{n_2} \alpha_{i'}^2 + \sum_{k=1}^{m_0} \beta_k^{(0)} + \sum_{j=1}^{m'_{12}} \beta_j^{(12)}.$$

Additionally, the conditions in Eq. (11)-(12) give rise to the constraints of

$$-\sum_{i=1}^{n_1} \alpha_i^{(1)} + \sum_{i'=1}^{n_2} \alpha_{i'}^{(2)} - \sum_{k=1}^{m_0} \beta_k^{(0)} + \sum_{j=1}^{m'_{12}} \beta_j^{(12)} = 0,$$

$$-\sum_{i=1}^{n_1} \alpha_i^{(1)} + \sum_{i'=1}^{n_2} \alpha_{i'}^{(2)} \geq 0.$$

The conditions in Eq. (13)-(16) give rise to the constraints of

$$0 \leq \alpha_i^{(1)} \leq C_1, i = 1,\ldots,n_1; 0 \leq \alpha_{i'}^{(2)} \leq C_1, i' = 1,\ldots,n_2,$$

$$0 \leq \beta_k^{(0)} \leq C_2, k = 1,\ldots,m_0; 0 \leq \beta_j^{(12)} \leq C_2, j = 1,\ldots,m'_{12}.$$

Finally, the dual problem becomes

$$\min_{\alpha,\beta} \frac{1}{2}\gamma^T Y K Y \gamma - \sum_{i=1}^{n_1} \alpha_i^{(1)} - \sum_{i'=1}^{n_2} \alpha_{i'}^{(2)} - \sum_{k=1}^{m_0} \beta_k^{(0)} - \sum_{j=1}^{m'_{12}} \beta_j^{(12)},$$

subject to

$$-\sum_{i=1}^{n_1} \alpha_i^{(1)} + \sum_{i'=1}^{n_2} \alpha_{i'}^{(2)} - \sum_{k=1}^{m_0} \beta_k^{(0)} + \sum_{j=1}^{m'_{12}} \beta_j^{(12)} = 0,$$

$$-\sum_{i=1}^{n_1} \alpha_i^{(1)} + \sum_{i'=1}^{n_2} \alpha_{i'}^{(2)} \geq 0,$$

$$0 \leq \alpha_i^{(1)} \leq C_1, i = 1, \ldots, n_1; 0 \leq \alpha_{i'}^{(2)} \leq C_1, i' = 1, \ldots, n_2,$$

$$0 \leq \beta_k^{(0)} \leq C_2, k = 1, \ldots, m_0; 0 \leq \beta_j^{(12)} \leq C_2, j = 1, \ldots, m'_{12}. \quad \blacksquare$$

## **MRI protocols, parametric maps, and image co-registration**

The MRI images used in this study were obtained through standard protocols and gone through preprocessing steps for quality control, which were described in detail in our previous publications [1]–[3]. Here we provide an exertion of the detailed approaches from a prior paper [1].

We performed all imaging at 3 T field strength (Sigma HDx; GE-Healthcare Waukesha Milwaukee; Ingenia, Philips Healthcare, Best, Netherlands; Magnetome Skyra; Siemens Healthcare, Erlangen Germany) within 1 day prior to stereotactic surgery. Conventional MRI included standard pre- and post-contrast T1-Weighted (T1-C, T1+C, respectively) and pre-contrast T2-Weighted (T2W) sequences. T1W images were acquired using spoiled gradient recalled-echo inversion-recovery prepped (SPGR-IR prepped) (TI/TR/TE = 300/6.8/2.8 ms; matrix = 320 × 224; FOV = 26 cm; thickness = 2 mm). T2W images were acquired using fast-spin-echo (FSE) (TR/TE = 5133/78 ms; matrix = 320 × 192; FOV = 26 cm; thickness = 2 mm). T1 + C images were acquired after completion of Dynamic Susceptibility-weighted Contrast-enhanced (DSC) Perfusion MRI (pMRI) following total Gd-DTPA (gadobenate dimeglumine) dosage of 0.15 mmol/kg as described below [2], [4], [5]. Diffusion Tensor (DTI): DTI imaging was performed using Spin-Echo Echo-planar imaging (EPI) [TR/TE 10,000/85.2 ms, matrix 256 × 256; FOV 30 cm, 3 mm slice, 30 directions, ASSET, B = 0,1000]. The original DTI image DICOM files were converted to a FSL recognized NIfTI file format, using MRIConvert (http://lcni.uoregon.edu/downloads/mriconvert), before processing in FSL from semi-automated script. DTI parametric maps were calculated using FSL (http://fsl.fmrib.ox.ac.uk/fsl/fslwiki/), to generate whole-brain maps of mean diffusivity (MD) and fractional anisotrophy (FA) based on previously published methods [6]. DSC-pMRI: prior to DSC acquisition, preload dose (PLD) of 0.1 mmol/kg was administered to minimize T1W leakage errors. After PLD, we employed Gradient-echo (GE) EPI [TR/TE/flip angle = 1500 ms/20 ms/60°, matrix 128 × 128, thickness 5 mm] for 3 min. At 45 s after the start of the DSC sequence, we administered another 0.05 mmol/kg i.v. bolus Gd-DTPA [2], [4], [5]. The initial source volume of images from the GE-EPI scan contained negative contrast enhancement (i.e., susceptibility effects from the PLD administration) and provided the MRI contrast labeled EPI+C. At approximately 6 min after the time of contrast injection, the T2*W signal loss on EPI+C provides information about tissue cell density from contrast distribution within the extravascular, extracellular space [2], [7]. We performed leakage correction and calculated relative cerebral blood (rCBV) based on the entire DSC acquisition using IB Neuro (Imaging Biometrics, LLC) as referenced [8], [9]. We also normalized rCBV values to contralateral normal appearing white matter as previously described [2], [5]. Image coregistration: for image coregistration, we employed tools from ITK (www.itk.org) and IB Suite (Imaging Biometrics, LLC) as previously described [2], [4], [5]. All datasets were coregistered to the relatively high quality DTI B0 anatomical image volume. This offered the additional advantage of minimizing potential distortion errors (from data resampling) that could preferentially impact the mathematically sensitive DTI metrics. Ultimately, the coregistered data exhibited in plane voxel resolution of ~ 1.17 mm (256 × 256 matrix) and slice thickness of 3 mm.

**Feature extracted from regional MRI**

The MRI features corresponding to each biopsy sample were extracted from a defined "region", i.e., an 8x8 pixel$^2$ window centered at the sampling location. From this window, we extracted 56 features from each of the five MRI contrast images, which included 18 statistical features and 26 and 12 texture features using two well-established texture analysis algorithms, Gray-Level Co-occurrence Matrix (GLCM) and Gabor

Filters (GF), respectively. The statistical features include commonly used ones in the literature [10], such as mean and standard deviation of gray-level intensities, Energy, Total Energy, Entropy, Minimum, 10th percentile, 90th percentile, Maximum, Median, Interquartile Range, Range, Mean Absolute Deviation (MAD), Robust Mean Absolute Deviation (rMAD), Root Mean Squared (RMS), Skewness, Kurtosis, Uniformity. Before applying GLCM and GF, we mapped the intensity values within the window onto the range of 0–255. This step helped standardize intensities and reduced effects of intensity nonuniformity on features extracted during subsequent texture analysis. The GLCM algorithm produced 26 features by setting the distance parameter to be 1 and 3 to capture different scales of spatial patterns [11], [12], such as Angular Second Moment Average, Contrast Average, Correlation Average, Sum of Squares Variance Average, Inverse Difference Moment Average, Average of Sum Average, Average of Sum Variance, Average of Sum Entropy, Entropy Average, Average of Difference Variance, Average of Difference Entropy, Average of Information Measure of Correlation 1, Average of Information Measure of Correlation 2. The GF algorithm produces 12 features, including Gabor Mean and Gabor Standard Deviation, by setting sigma to be 0.4, 0.7 and frequency to be 0.1, 0.3, 0.5 to capture different frequency and orientation contents [13].

**References**

[1] L. S. Hu *et al.*, "Uncertainty quantification in the radiogenomics modeling of EGFR amplification in glioblastoma," *Sci Rep*, vol. 11, no. 1, p. 3932, Feb. 2021, doi: 10.1038/s41598-021-83141-z.

[2] L. S. Hu *et al.*, "Multi-Parametric MRI and Texture Analysis to Visualize Spatial Histologic Heterogeneity and Tumor Extent in Glioblastoma," *PLoS One*, vol. 10, no. 11, p. e0141506, Nov. 2015, doi: 10.1371/journal.pone.0141506.

[3] N. Gaw *et al.*, "Integration of machine learning and mechanistic models accurately predicts variation in cell density of glioblastoma using multiparametric MRI," *Sci Rep*, vol. 9, no. 1, p. 10063, Jul. 2019, doi: 10.1038/s41598-019-46296-4.

[4] L. S. Hu *et al.*, "Radiogenomics to characterize regional genetic heterogeneity in glioblastoma," *Neuro Oncol*, vol. 19, no. 1, pp. 128–137, Jan. 2017, doi: 10.1093/neuonc/now135.

[5] L. S. Hu *et al.*, "Reevaluating the imaging definition of tumor progression: perfusion MRI quantifies recurrent glioblastoma tumor fraction, pseudoprogression, and radiation necrosis to predict survival," *Neuro Oncol*, vol. 14, no. 7, pp. 919–930, Jul. 2012, doi: 10.1093/neuonc/nos112.

[6] S. J. Price *et al.*, "Improved delineation of glioma margins and regions of infiltration with the use of diffusion tensor imaging: an image-guided biopsy study.," *AJNR Am J Neuroradiol*, vol. 27, no. 9, pp. 1969–74, Oct. 2006.

[7] N. B. Semmineh *et al.*, "Assessing tumor cytoarchitecture using multiecho DSC-MRI derived measures of the transverse relaxivity at tracer equilibrium (TRATE).," *Magn Reson Med*, vol. 74, no. 3, pp. 772–84, Sep. 2015, doi: 10.1002/mrm.25435.

[8] J. L. Boxerman, K. M. Schmainda, and R. M. Weisskoff, "Relative cerebral blood volume maps corrected for contrast agent extravasation significantly correlate with glioma tumor grade, whereas uncorrected maps do not.," *AJNR Am J Neuroradiol*, vol. 27, no. 4, pp. 859–67, Apr. 2006.

[9] L. S. Hu *et al.*, "Impact of Software Modeling on the Accuracy of Perfusion MRI in Glioma.," *AJNR Am J Neuroradiol*, vol. 36, no. 12, pp. 2242–9, Dec. 2015, doi: 10.3174/ajnr.A4451.

[10] A. Zwanenburg, S. Leger, M. Vallières, and S. Löck, "Image biomarker standardisation initiative," Dec. 2016, doi: 10.1148/radiol.2020191145.

[11] T. Ojala, M. Pietikainen, and T. Maenpaa, "Multiresolution gray-scale and rotation invariant texture classification with local binary patterns," *IEEE Trans Pattern Anal Mach Intell*, vol. 24, no. 7, pp. 971–987, Jul. 2002, doi: 10.1109/TPAMI.2002.1017623.

[12] R. M. Haralick, K. Shanmugam, and I. Dinstein, "Textural Features for Image Classification," *IEEE Trans Syst Man Cybern*, vol. SMC-3, no. 6, pp. 610–621, Nov. 1973, doi: 10.1109/TSMC.1973.4309314.

[13] A. G. Ramakrishnan, S. Kumar Raja, and H. V. Raghu Ram, "Neural network-based segmentation of textures using Gabor features," in *Proceedings of the 12th IEEE Workshop on Neural Networks for Signal Processing*, IEEE, pp. 365–374. doi: 10.1109/NNSP.2002.1030048.

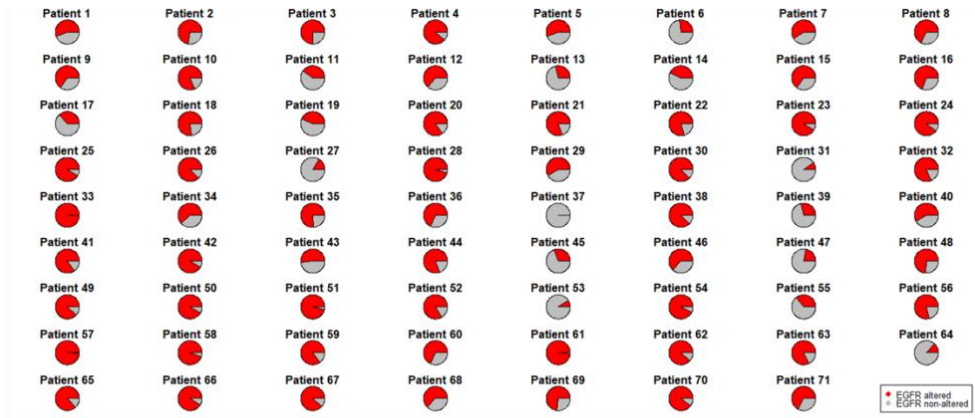

(a) EGFR

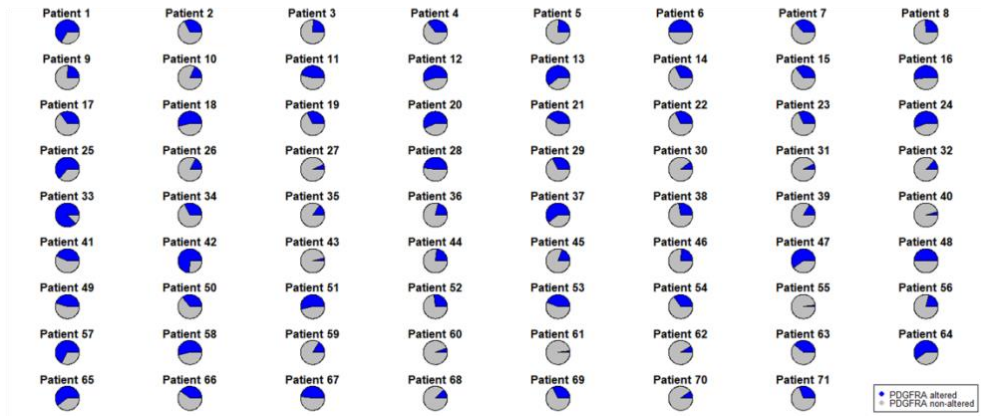

(b) PDGFRA

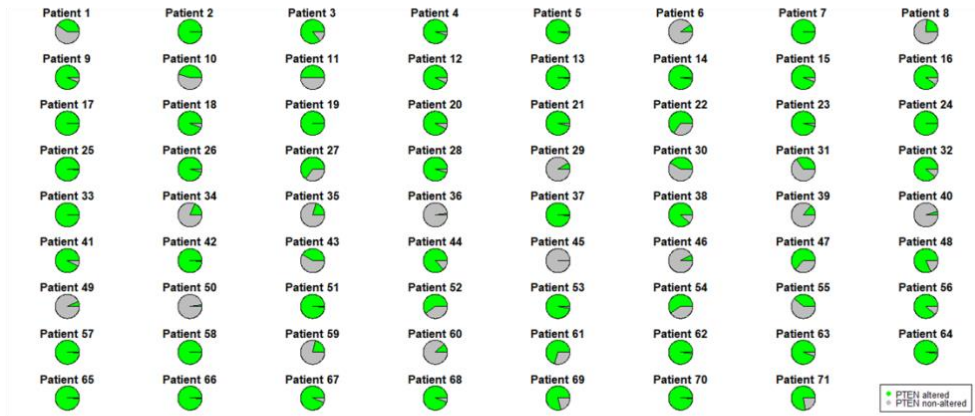

(c) PTEN

**S1 Fig. Patient-wise proportions of alteration vs. non-alteration for (a) EGFR, (b) PDGFRA, and (c) PTEN within tumoral AOI, aggregated from the prediction maps of these genes by WSO-SVM.**